\documentclass[sigconf, nonacm]{acmart}

\usepackage{xcolor}
\usepackage{listings}
\usepackage{subfigure}
\usepackage{multicol}
\usepackage{multirow}
\usepackage{enumitem}
\usepackage[linesnumbered,ruled,noend]{algorithm2e}

\newcommand{\oursys}{{\textsc{Angel-PTM}}}

\begin{document}
\title{\oursys{}: A Scalable and Economical Large-scale Pre-training System in Tencent}


\author{Xiaonan Nie}
\affiliation{%
  \institution{Peking University }
}
\email{xiaonan.nie@pku.edu.cn}

\author{Yi Liu}
\affiliation{%
  \institution{Tencent Inc.}
}
\email{callbackliu@tencent.com}

\author{Fangcheng Fu}
\affiliation{%
  \institution{Peking University }
}
\email{ccchengff@pku.edu.cn}

\author{Jinbao Xue}
\affiliation{%
  \institution{Tencent Inc.}
}
\email{jinbaoxue@tencent.com}

\author{Dian Jiao}
\affiliation{%
  \institution{Tencent Inc.}
}
\email{focusjiao@tencent.com}

\author{Xupeng Miao}
\affiliation{%
  \institution{Carnegie Mellon University}
}
\email{xupeng@cmu.edu}

\author{Yangyu Tao}
\affiliation{%
  \institution{Tencent Inc.}
}
\email{brucetao@tencent.com}

\author{Bin Cui}
\affiliation{%
  \institution{Peking University }
}
\email{bin.cui@pku.edu.cn}

\renewcommand{\shortauthors}{Xiaonan Nie, Yi Liu, Fangcheng Fu, Jinbao Xue, Dian Jiao, Xupeng Miao, Yangyu Tao, and Bin Cui}

\begin{abstract}
Recent years have witnessed the unprecedented achievements of large-scale pre-trained models, especially the Transformer models. Many products and services in Tencent Inc., such as WeChat, QQ, and Tencent Advertisement, have been opted in to gain the power of pre-trained models. In this work, we present \oursys{}, a productive deep learning system designed for pre-training and fine-tuning Transformer models. \oursys{} can train extremely large-scale models with hierarchical memory efficiently. The key designs of \oursys{} are the fine-grained memory management via the \emph{Page} abstraction and a unified scheduling method that coordinate the computations, data movements, and communications. 
Furthermore, \oursys{} supports extreme model scaling with SSD storage and implements the lock-free updating mechanism to address the SSD I/O bandwidth bottlenecks. 
Experimental results demonstrate that \oursys{} outperforms existing systems by up to $114.8\%$ in terms of maximum model scale as well as up to $88.9\%$ in terms of training throughput. 
Additionally, experiments on GPT3-175B and T5-MoE-1.2T models utilizing hundreds of GPUs verify the strong scalability of \oursys{}. 
\end{abstract}

\maketitle

\section{Introduction}
\label{sec:intro}
Large-scale pre-trained models, such as Transfomer models,  have achieved remarkable advancements in various fields such as computer vision~\cite{vit,liu2021swin,vmoe},
natural language processing~\cite{bert,gpt2,gpt3,t5},
speech recognition~\cite{dong2018speech_trans, wang2020speechtransformer}
and generative AI~\cite{chatgpt, stablediffusion} in recent years, outperforming traditional machine learning models and becoming the state-of-the-art approach. 
For example, Chat-GPT~\cite{chatgpt} is capable of generating human-like text and performing various NLP tasks with impressive accuracy and efficiency.
The success of Transformer models can be attributed to their ability to automatically learn and extract hierarchical representations of data, making them highly suited for complex tasks~\cite{foundationopportunities}. 

With the hope of achieving better performance, efforts are made to increase the scale of models --- the flagship NLP model size has been increasing at a rate of 240$\times$ for every 2 years~\cite{memorywall}, and ~\citet{scalinglaws} suggested that the trend of increasing model size will continue for better model quality. Such an explosive growth in model scale inevitably increases the computational and memory cost, and training large-scale Transformer models becomes extremely expensive. For instance, Microsoft totally adopts 4480 A100 80G GPUs for training Megatron-Turing NLG 530B~\cite{turing530b}, which would cost almost 70 million dollars for only purchasing these computing resources.
\nocite{DBLP:conf/icde/XieSLWGZDC22}

Undoubtedly, in order to simultaneously enable the superior ability of Transformer models in real-world applications and meet the rapid evolution of model scales, it is necessary for companies to re-think and re-design the productive deep learning systems in this era. Regarding the use cases and demands in Tencent Inc., we would like identify two key characteristics of deep learning systems.

\begin{itemize}[leftmargin=*]
\item \textbf{Easy-to-use and easy-to-scale}. 
In real-world productive applications, most users are deep learning researchers or data scientists that are good at designing task-specific model architectures. In contrast, they usually lack the expert knowledge or experiences in deploying and accelerating the model training process in the distributed manner. Consequently, the deep learning system should require only a few lines' modifications to parallelize the training tasks. Moreover, since productive clusters are multi-tenant by nature, the available hardware resources would be varying. Thus, we seek for seamless scalability. In other words, when users wish to tune the amount of resources for their tasks, there should be no need to re-configure their parallel schemes.

\item \textbf{Efficient and cost-effective}. 
Due to the stunning scale of models and datasets, training large-scale models is extremely time-consuming. Therefore, how to achieve a better training efficiency is vital to the practical deployment of pre-trained models. In addition to the running time, we should also improve the hardware usage ratio with best efforts. It is undoubtedly that it makes pre-trained models more economical if we can support the efficient training of large-scale models using as few resources as possible.
\end{itemize}

We observe that existing deep learning systems, such as Megatron-LM~\cite{megatron-lm} and DeepSpeed~\cite{rajbhandari2022deepspeed, zerooffload}, fail to achieve these characteristics. First, many systems involve complex parallelism strategies (such as tensor parallelism and pipeline parallelism) to scale the training tasks, which require the users to have a thorough understanding about how to split the models across the available accelerators (e.g., GPU devices) to optimize the distributed computing efficiency. This incurs extra efforts to deploy and scale the training tasks. Second, we find that existing systems adopt a coarse memory management to allocate and release the memory during training, which makes them unavoidably suffer from memory fragments and low resource usage ratio when training large-scale Transformer models, as we will analyzed more in-depth in Section~\ref{sec:usecases}.

In this work, we develop \oursys{}, a brand new deep learning system to support the booming applications of large-scale pre-training models in Tencent.
The main contributions of \oursys{} are summarized as follows:
\begin{itemize}[leftmargin=*]
    \item We analyze the characteristics and requirements of large-scale model training tasks in Tencent and propose the underlying designs of \oursys{} to address these requirements, which integrates data parallelism, parameter sharding, and hierarchical memory to gain the convenience of use and the transparency of scaling to various numbers of GPUs.
    \item To reduce the memory fragments and fully utilize the memory and bandwidth, 
    we propose the \textit{fine-grained {Page} abstraction} and manage the model states at the page level, including allocation, release, movement, and communication. Furthermore, we design the unified scheduler together with an \textit{fine-grained life-time based scheduling} method to dynamically manage these operations in a holistic manner for efficient training.
    \item To support enlarging models to an extreme scale, we integrate the SSD storage and design the \textit{Lock-Free Updating Mechanism} to eliminate the bottleneck of SSD I/O bandwidth.
    \item We have conducted evaluations on various representative large-scale Transformer models. Results show that \oursys{} achieves up to $114.8\%$ improvement in maximum supported model scale 
    and up to $88.9\%$ improvement in throughput performance 
    compared to existing systems. Experiments also verify the near-linear scalability of \oursys{} when training 
    on hundreds of GPUs. 
\end{itemize}

\oursys{} has been deployed in Tencent for around a year and supported a wide range of products and services, including WeChat, QQ, Tencent Games, Tencent Advertising, and Tencent Cloud, with its large models that span various AI domains such as NLP, CV, and cross-modal tasks. 
In addition, \oursys{} has also supported the training of HunYuan series large models, where the HunYuan-1T model achieved the first place in the overall ranking of the CLUE benchmark~\footnote{https://cluebenchmarks.com/rank.html}.

\section{Background}
\label{sec:background}
\subsection{Memory Management in Deep Learning}
As the size of the models increases, GPU memory management becomes a critical factor that affects the performance and scalability of deep learning models~\cite{gpt3, scalinglaws, nie2021evomoe}. In this section, we will provide a brief overview of the memory consumption during training and how they are managed in current deep learning frameworks.

\textbf{Deep Learning Training.} Deep learning models are composed of multiple layers of mathematical functions, with each layer taking the outputs of the previous layer as inputs and producing an output. 
Therefore, the computation of training can be represented as a computation graph, where each node stands for an operation, such as matrix multiply, and each edge is a tensor or dependency~\cite{abadi2016tensorflow}. 
The computation graph of a DNN is typically first converted into a list of operations using the topological sorting algorithm, and then the executor runs these operations in order.
To achieve a satisfactory model quality, the training process involves numerous forward and backward propagation passes. During the forward pass, a batch of training data is fed through the computation graph, utilizing the model parameters to produce activations for each operation. The final outputs are then compared to the expected values using a loss function.
During the backward pass, the error values are propagated back through the computation graph to compute the gradients of activations and parameters, respectively, where the gradients of parameters are further used by the optimizer to update the model parameters. 
In summary, the memory during training is primarily consumed by parameters and their gradients, activations and their gradients, and the optimizer states. Among them, the parameters and optimizer states will be preserved during training, while activations, gradients of activations, and gradients of parameters will be dynamically generated and released at each iteration. 
In the rest of this work, we would use the term ``model states'' to denote parameters and optimizer states for simplicity.

\begin{figure}[t]
    \centering
    \includegraphics[width=0.46\textwidth]{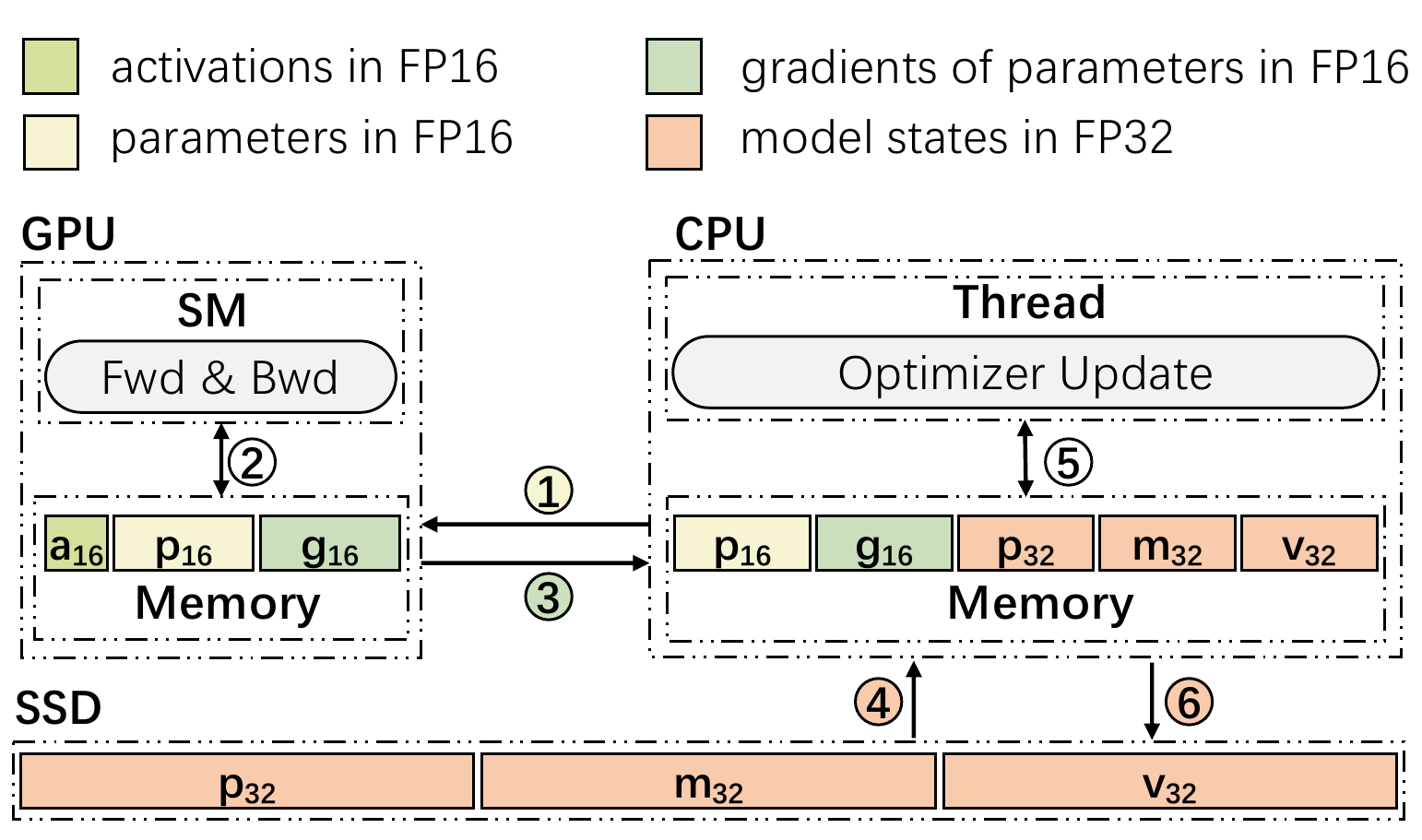}
    \caption{Illustration on the workflow of training on hierarchical memory.}
    \label{fig:hierarchical_training}
    \vspace{-3mm}
\end{figure}

\textbf{Mixed Precision Training.} To reduce the computation and memory requirements without sacrificing model quality, ~\citet{micikevicius2018mixed} proposed the \textit{mixed precision} techniques for training. As shown in Figure~\ref{fig:hierarchical_training}, the parameters are cast to the half-precision format (i.e., FP16 or the BF16 variant) before computation, so that activations and both types of gradients will be calculated in the FP16 fashion. Meanwhile, the model states are stored in the single-precision format (i.e., FP32) to preserve model quality. 
With the rapid growth of model size, \textit{mixed precision} has become a de facto paradigm for large-scale model training and deployment.

\textbf{Memory Management in Existing Frameworks.} 
Existing deep learning frameworks (e.g., PyTorch~\cite{paszke2019pytorch}, TensorFlow~\cite{abadi2016tensorflow} and Hetu~\cite{scis2022hetu}) employ a general memory management method, which manages the GPU memory in a separate memory pool that responds to the executor's requests, such as allocation and de-allocation.
For example, TensorFlow utilizes the best-fit allocation (BFC) algorithm to manage GPU memory and minimize memory fragmentation caused by frequent allocations and de-allocations. 
By allocating only the necessary amount of memory, this algorithm tries best to reduce the waste in GPU memory. Compared to other allocation algorithms, the BFC algorithm may take longer to find an available block, but it is well-suited for systems with limited GPU memory.

\textbf{Hierarchical Memory for Training.}
To accommodate the memory demands of training large models, 
many frameworks attempt to incorporate the hierarchical memory within GPU servers. 
To be specific, to address the memory consumption of Transformer-based pre-trained models, 
researchers have proposed shifting the optimizer states and computations to the CPU memory and the SSD storage~\cite{zeroinfinity, zerooffload, patrickstar}. 
We illustrate the workflow of training on hierarchical memory in Figure~\ref{fig:hierarchical_training}. The GPU (1) fetches the parameters from the CPU, (2) performs forward and backward computations on the GPU, and then (3) sends the calculated gradients back to the CPU.
The CPU (4) loads optimizer states from the SSD storage, (5) performs optimizer updating on CPU, and (6) stores the optimizer states on the SSD storage.

\begin{table}[t]
    \caption{Memory footprints of a single Transformer layer under the mixed-precision training with Adam optimizer.
    }
    \begin{center}
        \begin{tabular}{|c|c|c|c|c|}
        \hline
        Block & Layer & Params.(B) & Acts.(B) & \multicolumn{1}{c|}{Optims.(B)} \\
        \hline
        \hline
        \multirow{6}{*}{Attn} & Linear(Q,K,V) & $12d_{m}^2$ & $12{bsd_{m}}$ & $36d_{m}^2$ \\
        & MatMul & - & $4{bs}$ & - \\
        & {Scaled-} & \multirow{2}{*}{-} & \multirow{2}{*}{$4{bs}$} & \multirow{2}{*}{-} \\
        & {MaskSoftmax} & & & \\
        & MatMul & - & $4{bsd_{m}}$ & - \\
        & Linear & $4d_{m}^2$ & $4{bsd_{m}}$ & $12d_{m}^2$ \\
        \hline
        & Add & - & $4{bsd_{m}}$ & -  \\
        & LayerNorm & $4{d_{m}}$ & $4{bsd_{m}}$ & $12{d_{m}}$ \\
        \hline
        \multirow{3}{*}{FFN} & Linear & $4d_{m}d_{ffn}$ & $4{bsd_{ffn}}$ & $12d_{m}d_{ffn}$ \\
        & GeLU & - & $4{bsd_{ffn}}$ & - \\
        & Linear & $4d_{m}d_{ffn}$ & $4{bsd_{m}}$ & $12d_{m}d_{ffn}$ \\
        \hline
        & Add & - & $4{bsd_{m}}$ & -  \\
        & LayerNorm & $4{d_{m}}$ & $4{bsd_{m}}$ & $12{d_{m}}$ \\
        \hline
        \hline
        \multirow{2}{*}{Total} & & $16d_{m}^2$     &   $40bsd_{m}$    & $48d_{m}^2$\\
                               & & $+8d_{m}d_{ffn}$&   $+8bsd_{ffn}$    & $+24d_{m}d_{ffn}$\\
        \hline
        \end{tabular}
    \end{center}
\label{tab:footprints}
\end{table}

\subsection{Memory Footprints of Transformer}
A Transformer layer is stacked by a self-attention network and a position-wise feed-forward network (FFN), 
and it employs a residual connection on each of these two sub-layers, followed by a normalization layer~\cite{DBLP:journals/corr/layernorm}. 
In the following, we will approximately formulate the memory footprints of each component within a Transformer layer under the popular mixed precision training with the Adam optimizer scenario, where the input data is $X \in \mathbb{R}^{b \times s \times d_{m}}$. Specifically, $b$ is batch size, $s$ is sequence length, $d_{m}$ is hidden size of embeddings and $d_{ffn}$ is hidden size of FFN. Results of footprints are summarized in Table~\ref{tab:footprints}, and we ignore the small tensors for simplicity when calculating the total size, such as \textit{params.} of LayerNorm.

\textbf{Self-Attention.} The attention block~\cite{DBLP:conf/nips/VaswaniSPUJGKP17} could capture the dependencies between tokens in the sequence, and is effective in sequence modeling. As shown in Equation~\ref{equ:attention}, it first linearly projects the input $X$ into  queries ($Q$), keys ($K$) and values ($V$) with three linear functions respectively, where $\{W_{Q}, W_{K}, W_{V} \} \in \mathbb{R}^{d_{m} \times d_{m}}$ and the total footprint of \textit{Params} in this layer is $3 (\texttt{Q, K, V}) \times 2(\texttt{forward and backward}) \times 2 (\texttt{FP16, 2 bytes}) \times {d_{m} \times d_{m}} = 12d_{m}^2$.
Similarly, the \textit{Acts} is $\{Q, K, V\} \in \mathbb{R}^{b \times s \times d_{m}}$ and their footprint is $3 (\texttt{Q, K, V}) \times 2(\texttt{forward and backward}) \times 2 $
$ (\texttt{FP16, 2 bytes}) \times {b \times s \times d_{m}} = 12bsd_{m}$.
The footprint of model states (\textit{Optims}) is $3 (\texttt{Q, K, V}) \times 
3 (\texttt{master parameter, momentum, variance}) \times 4$ \newline
$(\texttt{FP32, 4 bytes}) \times {d_{m} \times d_{m}} = 12d_{m}^2$. 
The same calculation method is also applicable to other layers, we will directly give the results in the table~\ref{tab:footprints} because of the limited space.
After this linear layer, it will compute the attention scores (shape: $b \times s$) between each query-key ($Q$-$K$) pair, which is obtained by performing the dot-product (\texttt{MatMul}) operation as well as the \texttt{ScaledMaskSoftmax} operation. And the three operations, including \texttt{Scale}, \texttt{Mask(opt.)} and \texttt{Softmax}, are always fused into the \texttt{ScaledMaskSoftmax} operation for time-efficiency and memory-saving via kernel fusion techniques~\cite{wang2010kernel}.
The attention vectors (shape: $b \times s \times d_{m}$) are computed by weighted summation between the attention scores and values $V$. The self-attention layer finally produces the output (shape: $b \times s \times d_{m}$) by applying a linear transformation ($W \in \mathbb{R}^{d_{m} \times d_{m}}$) to the attention vectors.
\begin{align}
    \texttt{Attn}(X) &= \texttt{Softmax}\left(\texttt{Mask}\left( \frac{(XW_{Q})(XW_{K})^{T}}{\sqrt{d_k}}\right)\right)(XW_{V})
\label{equ:attention}
\end{align}

\textbf{Add \& LayerNorm.} 
A residual connection is employed between the input (shape: ${b \times s \times d_{m}}$) and output (shape: ${b \times s \times d_{m}}$) of the self-attention layer. Afterwards, a normalization layer is applied to the output of the Add operation (shape: ${b \times s \times d_{m}}$). These same transformations are also performed on the FFN block, and we formulate them in Equation~\ref{equ:norm_add}. Moreover, the size of parameters and their gradients of the LayerNorm layer is $4d_{m}$, one for weights and one for bias, which can be ignored compared to other parts.
\begin{align}
\label{equ:norm_add}
y = \texttt{LayerNorm}(f(x) + x), \quad where \ f \in \{ \texttt{Attention}, \texttt{FFN}\}
\end{align}

\textbf{Feed-Forward Networks.} As formualted in Equation~\ref{equ:ffn}, the feed-forward network (FFN) layer applies linear transformations to the inputs with two fully-connected (FC) layers separated by a GeLU activation function~\cite{hendrycks2016gelu}. 
Specifically, the first FC layer ($W_1 \in \mathbb{R}^{d_m \times d_{ffn}}$) projects the input into a new space with higher dimension, which allows the model to capture more complex relationships within a single token, 
while the second FC layer ($W_2 \in \mathbb{R}^{d_{ffn} \times d_{m}}$) shrinks the dimension back to original, which helps to ensure that the output of the model is well-behaved and that the overall architecture is not too complex.
\begin{equation}
    \texttt{FFN}(x_{s}) = W_{2} \cdot \texttt{GeLU}(W_{1} \cdot x_{s})
\label{equ:ffn}
\end{equation}

\textbf{Memory Usage Analysis.} According to Table~\ref{tab:footprints}, we can estimate the memory usage of any decoder-only Transformer models, where we do not take the \texttt{embedding\_look\_up} and \texttt{loss function} into consideration.
For the GPT-3 175B~\cite{gpt3}, the \textit{Params}, \textit{Acts} and \textit{Optims} consumes 648GB, 162GB, and 1944GB, respectively, when batch size ($b$) is 1, sequence length ($s$) is 2048, hidden size of embeddings ($d_m$) is 12288 and hidden size of FFN ($d_{ffn}$) is 49152. 
To satisify the memory requirement of large model training, multiple GPUs will be involved for distributed training with parallelism strategies. 

\subsection{Distributed Training}
By partitioning model parameters as well as their computation among multiple GPUs, the training can be significantly accelerated, enabling researchers to train large-scale Transformer models in a shorter amount of time. 
In this section, we will provide a comprehensive overview on existing parallelism strategies that have been widely adopted for Transformer models. 

\textbf{Data Parallelism and Zero Redundancy Optimization.} 
In data parallelism (DP), training samples are partitioned while model parameters and optimizer states are duplicated across multiple devices~\cite{limu2014scaling}. Each device executes the forward and backward propagation on its local mini-batch data to obtain its parameter gradients, and the gradients are averaged through a synchronization across all devices (e.g., by all-reduce). Eventually, each device updates the model parameters and optimizer states individually via the synchronized gradients. 
However, the vanilla data parallelism requires each device to maintain a full copy of model states, which is memory-inefficient for large-scale Transformer models. 
To reduce the memory consumption, 
~\citet{rajbhandari2020zero} proposed the Zero Redundancy Optimization (ZeRO) technique, which evenly partitions the model states across all devices. To be specific, when training with $N$ devices, each device only stores and updates $1/N$ of the model states. However, in each training iteration, an extra round of all-gather communication is needed to ensure each device gets the full updated parameters in order to accomplish the propagation. In short, ZeRO-powered data parallelism improves the memory efficiency at the cost of extra communication overheads.

\textbf{Model Parallelism and Hybrid Parallelism.} Model parallelism splits the model across multiple devices and performs the forward and backward propagation in a distributed manner.
Megatron-LM~\cite{megatron-lm} proposed tensor parallelism (TP), which partitioned 
the queries, keys and values matrices of the the attention network in a row- or column-parallel fashion, which exploits the inherent parallelism of the multi-head attention.  
In pipeline parallelism (PP), the model is partitioned into a sequence of stages and each stage is executed on a separate device.
~\citet{DBLP:conf/nips/Gpipe} proposed the batch-splitting pipeline algorithm and achieved almost linear speedup over multiple GPUs.
Hybrid parallelism refers to the combination of two or more parallelism strategies to improve the training efficiency, which must consider the three aspects of computation, communication, and storage simultaneously.
~\citet{miao2022galvatron} constructed a decision-tree based optimization approach to automatically find the optimal hybrid parallelism strategy for each layer of the model.

\section{Motivations and System Design}
\label{sec:usecases}
\subsection{Use Cases in Tencent}
Through collecting a significant amount of training task data from the largest machine learning platform in Tencent (i.e., Angel~\cite{jiang2018angel,zhang2019ps2,jiang2020psgraph}), we have identified two main categories of common use cases, including pre-training and fine-tuning. The majority of these tasks involve the large-scale training of Transformer models, as the Transformer architecture has revolutionized natural language processing and enabled efficient handling of sequential data with long-range dependencies. Each task category has its unique characteristics and primary objectives, and we provide a detailed analysis of these categories in the following section. Additionally, we briefly introduce corresponding system optimizations to improve their efficiency and effectiveness.

\textbf{Pre-Training.} Pre-training refers to the process of training a large-scale Transformer model on vast amounts of unlabelled data to learn rich and diverse features, which then can be useful for a wide range of downstream tasks, such as question answering and machine translation. 
Given the scale of models and datasets, pre-training tasks are extremely time-consuming and memory-hungry.
Meanwhile, ~\citet{scalinglaws} suggested that the model quality of pre-trained models scales as a power-law with data size, model size, and the amount of computation, which further increases the demand of pre-training tasks.

After analyzing the log information of the platform, we find that although pre-training tasks have to use hundreds or even thousands of GPUs to train for several weeks, 
they account for only about $10\%$ of the total number of tasks.
This is because researchers in Tencent prefer to jointly train a large-scale Transformer model as their shared base model. 
And we observe that there exist two main characteristics associated with pre-training:
\begin{itemize}[leftmargin=*]
    \item \textbf{Low-Efficiency on Scalability.} 
    During the running of pre-training tasks, users may pause and request more GPUs to obtain experimental results faster. However, in many cases the GPU utilization and training throughput decrease after more GPUs are involved, depending on the distributed training strategy of the model. 
    Take a real case as the example. Training a 64-layer GPT model with the hybrid parallelism strategy of Megatron-LM on 72 GPUs is slower than that on 64 GPUs. As discussed in Section~\ref{sec:intro}, many researchers in Tencent are not familiar with distributed computation and cannot adjust the parallelism strategy correspondingly.
    \item \textbf{Failure and Recovery.} 
    When more GPUs are involved, the Mean Time To Failure (MTTF) is shortened accordingly. Given the large amount of GPUs and the long training time, pre-training tasks would encounter GPU failure with a high probability, and should be restarted after failure.
\end{itemize}

\textbf{Fine-Tuning.} Fine-tuning refers to taking the pre-trained model and adapting it to a specific downstream task with domain-specific data, such as the supervised fine-tuning (SFT) phase in InstructGPT~\cite{ouyang2022instructgpt}. 
After fine-tuning the pre-trained model for a specific downstream task, researchers aim to deploy the resulting model in a real-world product. Therefore, they need to carefully tune hyper-parameters and iterate on experiments to achieve the best possible performance. This phase involves a trial-and-error progress and may require several iterations until a satisfactory model is obtained.

After analyzing the log information of the platform, we notice that the fine-tuning tasks account for about $90\%$ of the total number of tasks. Each fine-tuning task also requires a large number of GPUs, but the running time is shorter than pre-training tasks (usually in hours). 
And we find that there exists three main characteristics associated with fine-tuning:

\begin{itemize}[leftmargin=*]
    \item \textbf{Low-Efficiency on GPU Utilization.} Due to the small size of downstream datasets in fine-tuning tasks and to prevent overfitting, a small batch size is often used. However, this results in a significant amount of time spent on distributed communication, which reduces the utilization of expensive GPU computing units.
    \item \textbf{Long Response Time.} There are a large number of fine-tuning tasks in the task queue, each requiring certain amount of GPU resources. Due to the limited resources of the cluster, most tasks must wait, with waiting times up to several hours. Given that most fine-tuning tasks usually take merely a few hours, such a long response time severely hinders the development of productive applications.
\end{itemize}

\begin{table}[t]
    \small
    \caption{Distribution of tensor sizes within one layer of GPT3.}
    \vspace{-3mm}
    \begin{center}
        \begin{tabular}{|c||ccccc|}
        \hline 
        \textbf{Tensor Size (MB)} & 3072 & 2304 & 1152 & 768 & 576 \\ 
        \hline
        \textbf{Counts}& 4 & 6 & 4 & 20 &12 \\
        \hline
        \hline
        \textbf{Tensor Size (MB)} & 288 & 0.375 & 0.046875 & 0.0234375 &  \\ 
        \hline
        \textbf{Counts}& 8 & 4 & 6 & 4 & \\
        \hline
        \end{tabular}
    \end{center}
\label{tab:token_size_distribution}
 \vspace{-5mm}
\end{table}

\subsection{System Design}

Based on the above observation and analysis, we adopt the following three strategies as the underlying designs of our system to address the challenges. 

\textbf{Data Parallelism.} During the pre-training process, the number of GPUs required for training tasks is dynamic, which requires our distributed strategy to be easy-to-scale and have good scalability. Although model parallelism may have better throughput performance in some cases, the design of parallelism strategies requires expert knowledge, and it is difficult to migrate between different parallel degrees. Therefore, we determine to choose data parallelism as our basic distributed solution.

\textbf{Parameter Sharding.} In DP, each GPU needs to store the complete model states, which makes it difficult to support the training of large models. To address this, we adopt the parameter sharding approach proposed by ZeRO~\cite{rajbhandari2020zero}, which evenly splits each parameter among multiple GPUs. When a parameter needs to be calculated, the complete parameter is obtained through an all-gather operation. This approach significantly reduces the memory requirements of each GPU and allows us to train much larger models.

\textbf{Hierarchical Memory.} 
The issues of long response time and low resource utilization are significantly due to the large number of fine-tuning tasks as well as their excessive numbers of GPUs w.r.t. the relatively small batch sizes. 
To tackle these problems, we incorporate heterogeneous storage within GPU servers, thereby reducing the number of GPUs required for fine-tuning tasks. 
Additionally, leveraging heterogeneous storage also allows pre-training tasks to train larger models with the same numbers of GPUs.

After determining the underlying designs of our system, we find that simply combining these strategies still lead to insufficient resources utilization in existing systems~\cite{rajbhandari2022deepspeed}.
Therefore, we first analyze the distribution of tensor sizes of one Transformer layer in the GPT3-175B model during training, based on the formulation presented in Table~\ref{tab:footprints}. 
The results are summarized in Table~\ref{tab:token_size_distribution}, which demonstrates that tensor sizes vary greatly, from 3072MB to 0.02MB.
We identify that this would mainly bring two inefficiencies.

\textbf{Insufficient Memory Usage.} 
The model states are need to move between different storage spaces for training. However, due to the inconsistent sizes of tensors, the space cannot be directly reused. Instead, the allocator needs to be re-requested and released, resulting in memory fragments. As the training process continues and the model state is constantly moved, more and more memory fragmentation is generated, leading to inefficient memory usage.

\textbf{Insufficient Bandwidth Usage.} 
The model states need to be moved between the CPU and GPU through PCIe, or between GPUs through network communication. For large tensors, such as a 3072MB tensor in Table~\ref{tab:token_size_distribution}, there must be enough space in the GPU to start the communication. Prior to this, the communication bandwidth is unused, resulting in inefficient use of the bandwidth.

To cope with these inefficiencies, we first propose the \textit{{Page}-based memory organization} in Section~\ref{sec:page} to improve the memory usage and then design the \textit{Unified Scheduler} in ~\ref{sec:scheduler} to to fully utilize the bandwidth. Moreover, we also design a \textit{Lock-Free Updating Mechanism} to address the insufficiency in GPU usage when enlarging model to an extreme scale.

\begin{figure}
    \centering
    \includegraphics[width=0.65\linewidth]{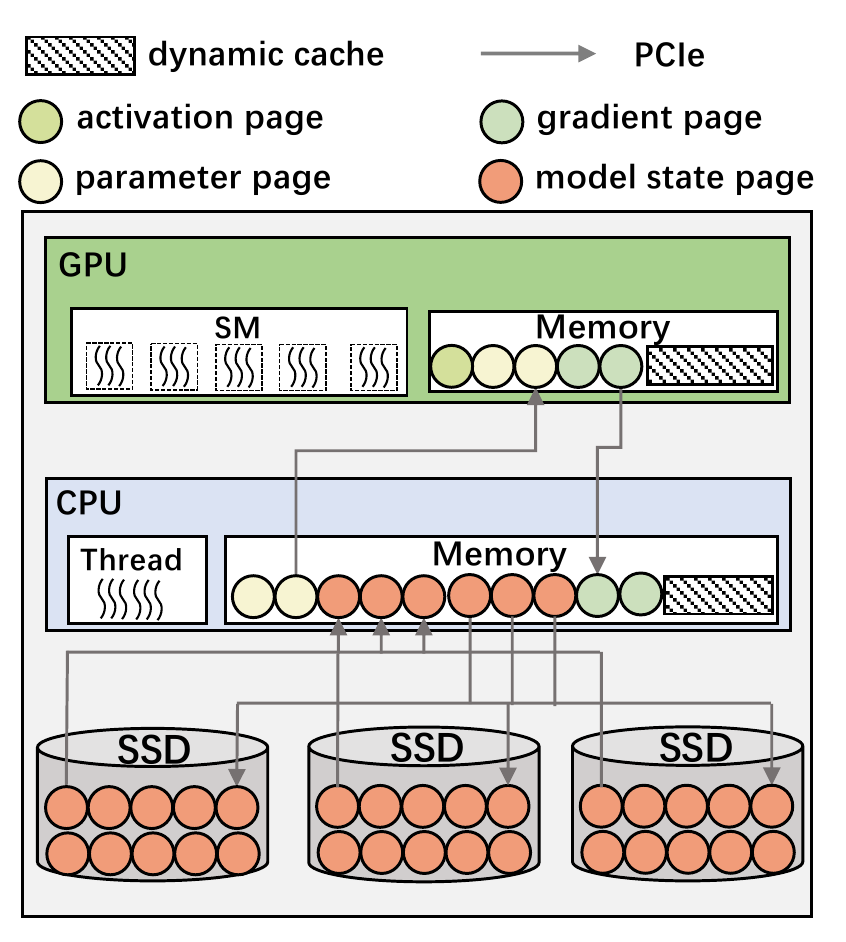}
    \caption{System architecture of \oursys{}.}
    \label{fig:overview}
\end{figure}

\section{\oursys}
\label{sec:design}
Our system, namely \oursys{}, is designed for researchers and developers to design and experiment large-scale Transformer models in Tencent. 
Figure~\ref{fig:overview} illustrates the system architecture, where we enable the fine-grained memory management at the \emph{Page} level and adopt the dynamic cache techniques for sufficient memory usage.
In the following subsections, we will analyze the limitations of existing systems and present our solutions to address these issues. 

\subsection{Page-Based Memory Organization}
\label{sec:page}
\textbf{Insufficient Memory Usage.} Existing systems suffer from memory fragments due to their coarse memory management. For instance, DeepSpeed uses the original memory management of PyTorch for offloading and recomputing, which frequently allocates and releases tensors, leading to space fragments because the sizes of these tensors are not uniform as discussed in Section~\ref{sec:usecases}.
PatrickStar~\cite{patrickstar} manages GPU memory in chunks rather than tensors, where the chunk size must be larger than the largest tensor used in model training. This would also result in memory fragments within each chunk as well as the in-efficiency of the overlapping between communication and computation. 

To reduce the fragments of memory space organization and improve the efficiency of memory movements, we propose the \emph{Page} abstraction, which works as the minimum unit of memory operations for heterogeneous storage, including allocation, release, movement, and remote communication. Each tensor in the model states is composed of several pages.

\setlength{\textfloatsep}{0.6cm}
\begin{figure}[t]
	\centering
	\begin{minipage}[t]{0.94\linewidth}
	\begin{lstlisting}[morekeywords={Tensor, vector, bool, struct}]
struct Page {
  /* Page Information */
  void*  data_ptr;
  size_t total_bytes;
  size_t available_bytes;
  // device_map: {0: GPU, 1: CPU, 2: SSD}
  size_t device_index;
  // ids for tensors in this page
  size_t tensor_id[2];
  // occupied bytes for each tensor
  size_t tensor_bytes[2];
  
  /* Page Interface */
  // allocate required bytes for id-th Tensor
  void allocate(size_t required_bytes, size_t id);
  // release space of id-th Tensor
  void release(size_t id);
  // move this page to target device asynchronously
  void move(size_t target_device_index);
  // send this page to id-th server asynchronously
  void send(size_t id);
  // receive contents from id-th server
  void receive(size_t id);
};

	\end{lstlisting} \vspace{-1ex}
	\end{minipage}
	\caption{The Page Abstraction.}
	\label{fig:page}
	\vspace{-10pt}
\end{figure}

\textbf{\emph{Page} Abstraction.} 
As illustrated in Figure~\ref{fig:page}, the \emph{Page} abstraction includes several key pieces of information, including a pointer to the actual memory data, the total number of bytes in the page, the number of bytes that are available for the next allocation, and the index of the device where the page is currently located. 
Additionally, each \emph{Page} can be associated with one or more tensors, with unique identifiers and information about the amount of memory occupied by each tensor.
In order to simplify the memory management, we decide to limit each page to contain information about a maximum of two tensors at any given time.
The reasoning behind this decision will be analyzed in conjunction with the discussion of the optimal page size in detail below.

Moreover, \emph{Page} also provides several interfaces for accessing and manipulating the data stored within it. These interfaces enable developers to perform a wide range of operations on \emph{Page} objects, including allocating and releasing memory for specific tensors, moving pages between the heterogeneous memory, and sending/receiving pages across different servers.

\textbf{Optimal \emph{Page} Size.} 
The optimal size of the \emph{Page} structure is a critical factor in balancing the competing demands for memory management efficiency and overall throughput.
On the one hand, if the \emph{Page} size is too large, there will be a large number of tensors coexisting in the page, increasing the complexity of our management and resulting in wasted space.
On the other hand, if the \emph{Page} size is too small, there will be increased overhead associated with data movement because of the under-utilized bandwidth.
Therefore, we believe that the minimum \emph{Page} size that can fully utilize the PCIe bandwidth is optimal for our system, i.e., 4MB. 

In fact, according to our observations during training, the vast majority of model states are larger than 4MB, which is also verified in Table~\ref{tab:footprints}.
Therefore, by carefully arranging these tensors, we can ensure that each page is associated with at most two tensors, which largely reduces the complexity of management.
For tensors that are smaller than 4MB, we allow each of them to occupy an individual page for simplicity, considering that they only account for a very small fraction of the overall memory usage.

\textbf{\emph{Tensor} Management.}
The \emph{Tensor} structure is a fundamental data structure in our system that represents multi-dimensional arrays of numerical data, composed of at least one page. As shown in Figure~\ref{fig:tensor}, it contains several crucial pieces of information, such as a unique id associated with the tensor, data type, shape, and the current device index.\footnote{
We set the device index as -1 when the tensor is not ready for computation (i.e., some of its pages need to be fetched from heterogenous memory or other servers).}
The \emph{Tensor} structure also provides a set of interfaces for memory management, such as allocating a certain shape tensor or releasing its data. Additionally, it offers an explicit interface for moving data between heterogeneous memory. Since the space of different pages may not be contiguous, the \emph{Tensor} structure provides the merge interface to make them contiguous.

\setlength{\textfloatsep}{0.6cm}
\begin{figure}[t]
	\centering
	\begin{minipage}[t]{0.93\linewidth}
	\begin{lstlisting}[morekeywords={vector, bool, struct}]
struct Tensor {
  /* Tensor Information */
  size_t  id;
  vector<Page> page_list;
  size_t  dtype;
  size_t* shape;
  size_t  device_index;
  
  /* Tensor Interface */
  void allocate(size_t* shape, size_t dtype);
  void release();
  void move(size_t target_device_index);
  void merge();
};
	\end{lstlisting} \vspace{-1ex}
	\end{minipage}
	\caption{The Tensor structure in our system.}
	\label{fig:tensor}
	\vspace{-10pt}
\end{figure}

\subsection{Unified Scheduler}
\label{sec:scheduler}
\textbf{Insufficient Bandwidth Usage.} 
As model states are transferred across heterogeneous memory within a server over the PCIe communication as well as between different servers over the network communication, this would result in significant time overheads and present a challenge for designing an effective communication scheme.   
How to make the best use of PCIe and communication bandwidth is crucial to achieving high throughput. To this end, it is necessary to \textit{minimize the amount of memory swapping} and \textit{maximize the overlap between communication and computation}. 
Existing systems, such as DeepSpeed, employ static data partitioning and data transfer scheduling, which can impede overall training speed. This is because, even when the GPU has sufficient memory, these systems still transfer the entire optimizer states and the update operations to the CPU, causing unnecessary data movements.

\begin{figure}[t]
    \centering
    \includegraphics[width=0.8\linewidth]{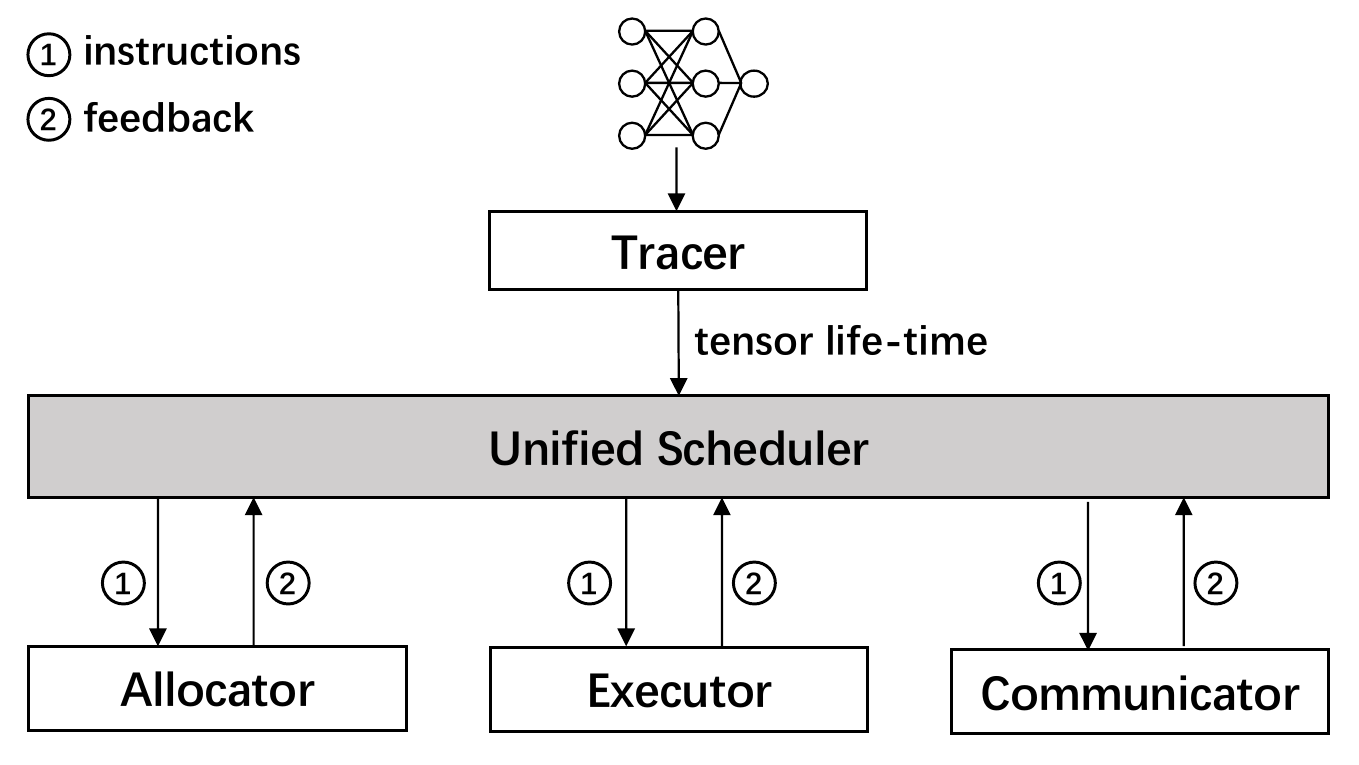}
    \caption{Illustration of the Unified Scheduler and associated components in \oursys{}.}
    \label{fig:workflow}
    \vspace{-5mm}
\end{figure}

To efficiently utilize the hierarchical and complicated network resources within GPU servers, we design a unified scheduling method on top of the \emph{Page} abstraction for efficient training.
As shown in Figure~\ref{fig:workflow}, once the model is defined by the user, the Tracer obtains the tensor access pattern and the life-time for each tensor (detailed in Section~\ref{sec:impl}). Then, the Unified Scheduler takes these statistics as input and schedules each operation at the right time during training.
The essential idea is that training of deep learning models is iterative by nature, so we can make a schedule on the operators given the tensor access pattern. 
To be specific, by analyzing the life-time of each tensor as well as its size, the Unified Scheduler deduces the most appropriate time to launch each operation, including calling the Allocator to move tensors, calling the Executor to perform GPU computations, and calling the Communicator for inter-GPU communication. 

\textbf{Tensor Life-Time.} The life-time of a tensor refers to the duration from its first access time to its last access time within a training iteration.
The key characteristic of deep learning training is the iterative nature, which means computation operations are inserted into the computing stream sequentially and iteratively. 
This allows us to optimize the sequence of data movements and communications with respect to computation, and ensures that the required data is available at the right time for each computation, reducing idle time and improving memory efficiency.
Additionally, since computation is performed at the tensor level, tensor allocation and release can also be done at this level. Using logical IDs instead of real-time for lifetime tracking simplifies the scheduling process.
By utilizing tensor lifetime information, 
we can optimize the scheduling of computations, movements, and communications in a holistic manner. 

\textbf{Unified Scheduler.} 
The Unified Scheduler is the crucial component in \oursys{}, responsible for coordinating the activities of other components, such as Allocator, Executor and Communicator. 
Specifically, it receives the life-time information and the access pattern from the Tracer and dynamically manages the memory, computation, and communication, which achieves high training throughput by maximizing the overlapping of different resources and reducing unnecessary operations.

First, we must determine an effective way to distribute the model states to various memory devices, such as GPU memory, CPU memory and SSD storage, and distribute computation tasks to various computing devices, such as CPUs and GPUs, which is a complex NP planning problem and cannot be solved directly for large-scale models. 
To tackle this problem, we develop a heuristic method and incorporate empirical information to aid in the design of our strategies. To be specific, the forward and backward computations of the transformer models are mainly composed of FP16 matrix multiplication, which is rather compute-intensive and requires less memory. The optimizer update computations, to the contrary, are composed of FP32 matrix addition, which is memory-intensive and takes less time to compute. Given the fact that GPUs have high computing capabilities but limited memory, we prioritize putting the forward and backward computations into the GPU, and the optimizer update computations into the CPU. The SSD storage will also be involved to store the FP32 optimizer states for scaling model size.
With this approach, it is necessary to transfer the FP16 parameters from the CPU to the GPU, and subsequently transfer the FP16 gradients from the CPU back to the GPU. These data movements need to be carefully scheduled to overlap with computations and avoid introducing extra overhead.
Meanwhile, we utilize the recomputation technique~\cite{chen2016recompute} to further alleviate the GPU memory pressure, where some activations are released in the forward pass and then are regenerated in the backward pass by re-executing their forward computation.

Second, we leverage caching techniques to fully exploit the high-speed memory and computation capabilities of the GPUs after previous assignments. For instance, if sufficient space is available, we reserve a portion of the GPU memory as the cache to store a segment of the CPU's optimizer states. Additionally, we move the relevant CPU computations to the GPUs, which reduces memory transfers and accelerates computation, leading to an overall improvement in training speed. 
Moreover, we dynamically make cache size decisions for each model based on its tensor lifetime information, ensuring training without encountering GPU out-of-memory errors. 
For medium-scale models, we can store and compute a large portion of tensors on the GPUs. Such a caching technique allows us to maximize GPU memory utilization and minimize data transfer overheads.

\begin{algorithm}[t]
    \small
	\SetAlgoLined
	\SetKwProg{Fn}{Function}{}{end}
	\SetKwInput{KwData}{Input}
	\SetKwInput{KwResult}{Ouptut}
    \SetKwFunction{myFunc}{Rollback}
    \SetKwProg{Fn}{Function}{:}{}
	\KwData{$model$: List of layers, where $l_i$ is $i$-$th$ layer\\
    \quad \quad \quad $traces$: List of trace for each tensor\\
    \quad \quad \quad $gpuMemory: $ total GPU memory \\
    }
	\KwResult{$S$: List of tasks, each is \{operation, page, trigger\_id \}}
	\SetInd{0.61em}{0.61em}
    /* Phase 1: Prioritize \texttt{move\_to\_gpu} tasks */\\
    $wait\_stack$ = \{ \}\;
    \For{$l_i$ $\in$ $M$}{
      \For{$page$ $\in$ $l_i.param.page\_list$}{
       $S$.append(\{\texttt{move\_to\_gpu}, $page$, $0$\})\;
      }
    }
    \For{$l_i$ $\in$ $M$}{
        \While{\texttt{get\_available\_memory}($S$, $traces$) < \text{size($l_i$)}}{
            $task \gets $ pop the last movement task from $S$\;
            $wait\_stack$.push\_back($task.page$)\;
        } 
        \For{$page$ $\in$ $l_i.param.page\_list$}{
           $S$.append(\{\texttt{all\_gather}, $page$, $i$\})\;
        }
        $S$.append(\{\texttt{compute}, \text{$l_i$}, $i$\})\;
        \While{$wait\_stack$\textup{.non\_empty()} and \texttt{get\_available\_memory}($S$,  $traces$) > \text{page\_size}}{
            $page \gets wait\_stack$.pop\_back()\;
            $S$.append(\{\texttt{move\_to\_gpu}, $page$, $i$\})\;
        }
    }
    ~\\
    /* Phase 2: Advance \texttt{all\_gather} tasks to overlap them with previous computation if no out-of-memory (OOM) */ \\
    \For{$task$ $\in$ $\{t | t \in S, t.operation == \texttt{all\_gather} \}$}{
        $task$ $\gets$ pop $task$ from $S$\;
        $id$ $\gets$ get the earliest possible id to trigger $task$ without OOM\;
        $S$ $\gets$ insert \{\texttt{all\_gather}, $task.page$, $id$ \} into $S$\;
    }
    \KwRet{$S$}\; 
\caption{Fine-grained Life-time based Scheduling}
\label{alg:scheduling}
\end{algorithm}

As presented in Algorithm~\ref{alg:scheduling}, the scheduling method takes the input model, the trace information for each tensor, and the total GPU memory as inputs to generate a task schedule that coordinates the other components of the system.
The algorithm consists of two phases.
In phase 1, we prioritize the movement of data to the GPU by inserting \texttt{move\_to\_gpu} tasks for each page of each tensor at the beginning of the schedule (lines 3-5). 
This is based on our prior knowledge that the speed of CPU-GPU data transfer (32GB/s) is slower than that of GPU-GPU communication (200GB/s). 
We then iterate over each layer to insert \texttt{all\_gather} tasks and \texttt{compute} tasks on demand (lines 10-12). 
If the current available memory is not sufficient for the running of the current layer, we will pop the last movement task until the memory requirement is met (lines 7-9).
If the current available memory is sufficient for the movement of the next page, we will schedule it immediately (lines 13-15).
In phase 2, we advance \texttt{all\_gather} tasks as early as possible to overlap them with previous computation tasks. 
Specifically, for each \texttt{all\_gather} task, we gradually try to shift its scheduled time earlier (i.e., decrease its trigger id by 1). In each trial, the trace information, tensor sizes, and the schedules after shifting will be utilized to measure the peak memory usage in order to determined whether there will be an out-of-memory (OOM) error after the shifting. And this progress ends at the smallest trigger id that does not encounter OOM error. Eventually, the \texttt{all\_gather} task will be inserted back to the schedule, along with the new trigger id (Line 20-21).
This approach contributes to more efficient use of available memory and better overlapping of tasks on the GPU, ultimately leading to improved performance.

\setlength{\textfloatsep}{0.1cm}
\begin{algorithm}[t]
    \small
	\SetAlgoLined
	\SetKwProg{Fn}{Function}{}{end}
	\SetKwInput{KwData}{Input}
	\SetKwInput{KwResult}{Ouptut}
	\KwData{
        $model = \{l_0, ..., l_{n-1}\}$:  list of layers, where $l_i$ is $i$-th layer\\
            \quad \quad \quad $p_{32}(l_i)$:  parameters for $l_i$ in FP32\\
            \quad \quad \quad $m_{32}(l_i)$:  first moments of gradients for $l_i$ in FP32\\
            \quad \quad \quad $v_{32}(l_i)$:  second moments of gradients for $l_i$ in FP32\\
            \quad \quad \quad $g_{16}(l_i)$:  gradients for $l_i$ in FP16\\
            \quad \quad \quad $p^\prime_{16}(l_i)$:  buffered parameters for $l_i$ in FP16\\
            \quad \quad \quad $g^\prime_{16}(l_i)$:  buffered gradients for $l_i$ in FP16\\
            }
	\SetInd{0.61em}{0.61em}
    /* Updating thread on CPU */ \\
    \While{there are uncleared buffered gradients}{
    \For{$l_i \in \texttt{reverse}(model)$}{
        \text{Fetch $p_{32}(l_i), m_{32}(l_i), v_{32}(l_i)$ from SSD storage}\;
        Update $p_{32}(l_i), m_{32}(l_i), v_{32}(l_i)$ via $g^\prime_{16}(l_i)$\;
        Pass $p_{32}(l_i)$ to the buffering thread\;
        Offload $p_{32}(l_i), m_{32}(l_i), v_{32}(l_i)$ to SSD storage\;
    }
    }
    ~\\
    /* Buffering thread on CPU */ \\
    \While{not finished}{
    \If{received $p_{32}(l_i)$ from the updating thread}{
        Clear buffered gradients: $g^\prime_{16}(l_i) \gets 0 $\;
        Update buffered parameters: $p^\prime_{16}(l_i) \gets \textup{cast}(p_{32}(l_i), \textup{FP16})$\;
    }\uElseIf{received $g_{16}(l_i)$ from the GPU}{
        Accumulate buffered gradients: $g^\prime_{16}(l_i) \gets g^\prime_{16}(l_i) + g_{16}(l_i)$\;
    }
    }
    ~\\
    /* Computation on GPU */ \\
    \While{not finished}{
    \For{$l_i \in model$}{
        \text{Fetch buffered parameters $p^\prime_{16}(l_i)$ from CPU memory}\;
        Perform forward computation with $p^\prime_{16}(l_i)$\;
    }
    \For{$l_i \in \texttt{reverse}(model)$}{
        Perform backward computation with $p^\prime_{16}(l_i)$ and get $g_{16}(l_i)$\;
        Offload $g_{16}(l_i)$ to CPU memory (inform the buffering thread)\;
    }
    }
    
\caption{Lock-Free Updating Mechanism}
\label{alg:async_training}
\end{algorithm}

\subsection{Lock-Free Updating Mechanism}
\label{sec:lock_free}
\textbf{Insufficient GPU Utilization.} 
When involving heterogeneous memory or multiple GPUs into training, the computation of GPUs is often blocked by the data transfer. For one thing, GPUs have to wait for updated parameters to be moved in for computations, or computed gradients to be moved out to make room. For another, GPUs also have to wait for gradient synchronization via remote communication to update local parameters. Undoubtedly, it results in the waste of GPU resources and reduces the overall efficiency.
Worst still, incorporating SSD storage into training further decreases GPU utilization in existing systems, as SSDs have the slowest I/O speed.
Take the A100 Server in Tencent as an example. The I/O speeds for GPU memory access, CPU-GPU transfer, and SSD-CPU transfer are 600 GB/s, 32 GB/s, and 3.5 GB/s, respectively. Our observations reveal that after introducing CPU memory and SSD storage, nearly $80\%$ of the iteration time is idle, whereas the number is merely $10\%$ when introducing only CPU memory.

To mitigate this training bottleneck, we design a novel \textit{Lock-Free Updating Mechanism}, 
which decouples the GPU computation from the CPU optimizer operations through a novel asynchronous consistency control protocol.
Algorithm~\ref{alg:async_training} illustrates the details of our proposed mechanism.
The essential idea is to employ two buffers in CPU memory to store the FP16 parameters and gradients respectively, and leverage an auxiliary buffering thread to maintain the buffers. 
During training, each GPU fetches the FP16 parameters from the CPU buffer and perform forward and backward computations (Line 19-23). Then, the generated gradients are offloaded to the CPU memory (Line 24), and eventually accumulated into the CPU buffer by the buffering thread (Line 15).
During optimizer updating, the updating thread on CPU reads the FP32 parameters and optimizer states from SSD, which are then updated according to the buffered, accumulated gradients (Line 4-5).
Subsequently, the buffering thread will be informed to update the buffers, including clearing the buffered gradients and casting the updated parameters into the buffered parameters (Line 12-13), which are overlapped with the offloading of the updated parameters and optimizer states to SSD (Line 7).

By introducing the auxiliary buffers, our system gets rid of the locks between GPU computations and CPU model updating. Although it incurs extra memory overhead, it is acceptable since both the buffered parameters and gradients are stored in FP16, requiring small memory footprints. Another side effect of the lock-free mechanism is that it may introduce staleness into the parameters given the fact that the updating thread on CPU runs slower than the GPU due to the limited SSD I/O bandwidth. Nevertheless, existing studies have verified that deep learning model training can well tolerate such staleness. In Section~\ref{sec:experiment}, we will empirically show that the convergence is not harmed while enjoying the efficiency improvement brought by the lock-free mechanism.
Last but not least, since the optimizer updates parameters element-wisely, the data movement and CPU computations can be scheduled at the \emph{Page} level, which better overlaps different resources.

\section{Implementation}
\label{sec:impl}
\setlength{\textfloatsep}{0.6cm}
\begin{figure}[t]
	\centering
	\begin{minipage}[t]{0.93\linewidth}
	\begin{lstlisting}[morekeywords={class, def, in, bool}]
import angelptm

class MyModel():
    def __init__(self)
        ...
    def forward(self, x)
        ...

model = MyModel()
optimizer = Optimizer(model)
model = angelptm.initialize(model, 
                            optimizer, config)
...
for batch in batches:
    loss = model(batch)
    model.backward(loss)
    model.step()

	\end{lstlisting} \vspace{-1ex}
	\end{minipage}
	\caption{Programming Interface.}
	\label{fig:interface}
\end{figure}

\oursys{} offers a comprehensive solution for efficient deep learning model training in industrial settings. 
It leverages some key techniques~\cite{miao2021het, DBLP:conf/sigmod/MiaoNSYJM021,nie2022tsplit} from Hetu~\cite{scis2022hetu}, gets implemented over PyTorch~\cite{paszke2019pytorch}, and features the \emph{Page} abstraction for memory efficiency and a unified scheduling method for resource utilization.
Furthermore, \oursys{} has undergone extensive optimization on A100 servers, enabling it to take full advantage of hardware capabilities for deep learning tasks. 
We would like to briefly go through the implementation of the key components in this section.

\textbf{Tracer.} 
The Tracer in \oursys{} is responsible for tracking the usage of each tensor and summarizing a tensor access pattern for the given model as a list of following elements:
\begin{itemize}[leftmargin=*]
    \item $tensor\_id$: The logical ID of this tensor.
    \item $first\_id$: The logical ID when first accessing this tensor.
    \item $end\_id$: The logical ID when last accessing this tensor.
    \item $cpu\_time$: The time for producing this tensor on CPU.
    \item $gpu\_time$: The time for producing this tensor on GPU.
\end{itemize}
To assign a unique tensor ID to each parameter, we modify the \textit{\_\_init\_\_} method of the Parameter class to use a global variable \textit{id}. Then, we track the first and last use of each parameter during an iteration by registering \texttt{hook} functions, recording them as \textit{first\_id} and \textit{end\_id}, respectively.
To capture the generation time of tensors on both CPUs and GPUs, we use the \texttt{time.time()} and \texttt{CudaEvent} interfaces respectively to accurately measure the CPU and GPU time for each tensor.

\textbf{Unified Scheduler.} 
The Unified Scheduler is responsible for coordinating the activities of three components, which are the Allocator, Executor, and Communicator.
Sending instructions by the message passing will bring severe overheads into training, thus we adopt the event-driven programming techniques. For example, computations will be launched into threads only if the events of modifying its input tensor are completed.

\textbf{Allocator.} 
The Allocator in \oursys{} is responsible for managing tensors at the \emph{Page} level in the hierarchical memory resources. It provides three memory operations for each page, including \texttt{allocate}, \texttt{release} and \texttt{move}. 
To reduce the overhead of requesting memory space and take advantages of the iterative nature of training, we pre-allocate space from the hierarchical memory of the system, including GPU memory, CPU pinned memory, and SSD memory.
To enable fine-grained memory operations, we divide the pre-allocated memory into pages of fixed size, where each page can be allocated, released and moved independently. Moreover, we utilize \texttt{cudaMemcpyAsync()} and the \texttt{DeepNVMe} library~\cite{zeroinfinity} for asynchronous GPU-CPU and CPU-SSD data movements, respectively.

\textbf{Executor.} The Executor in \oursys{} is responsible for scheduling the computation of Tensors on computational devices such as CPUs and GPUs on the server.
Meanwhile, it maintains a separate stream for each of these computational devices, including a CPU stream and a GPU stream.
By receiving instructions from the unified scheduler, it inserts computations into the corresponding stream and schedules them to the computation threads in the order of insertion. 
When all the inputs for the computation are ready, the computation begins, and feedback is sent back to the unified scheduler after the computation is complete.
The Executor in \oursys{} plays a critical role in the \textit{dynamic computation management mechanism}, which optimizes the overlap between the uses of different computational resources.

\textbf{Communicator.} 
The Communicator in \oursys{} is responsible for scheduling communication between different network devices, including NIC and NVLink. 
We implement the Communicator by using the NCCL library~\cite{jeaugey2017nccl}, which provides efficient communication primitives for multi-GPU systems. These primitives include collective operations such as \texttt{AllReduce}, \texttt{AllGather}, and \texttt{ReduceScatter}, which are essential for exchanging model parameters and gradients between GPUs during training. The Communicator also maintains a queue to store communication tasks and schedules them for execution based on instructions from the Unified Scheduler, thus it enables reordering the tasks in the queue to improve the overlap between computation and communication.

\textbf{Efficient  Movement on Distributed Servers.} 
GPU servers typically have a complex interconnect topology, such as A100 servers~\cite{a100} that contain two CPUs, four PCIe switches, and eight A100 GPUs. These GPUs can communicate with the CPU memory in parallel, providing efficient data movement in distributed training.
To take full advantage of this hardware feature, we evenly partition the model parameters across GPUs to parallelize the movement of parameters between the CPU and GPUs, which is similar to ZeRO-Infinity~\cite{zeroinfinity}. 
This further accelerates data movement during training and achieves excellent scalability.

\begin{table}[t]
    \caption{Overview of hardware environments.}
    \vspace{-3mm}
    \begin{center}
        \begin{tabular}{cccccc}
        \hline 
        \multicolumn{6}{|c|}{The Configuration of one GPU server} \\ 
        \hline
        \hline
        \multicolumn{2}{|c|}{CPU} & \multicolumn{4}{c|}{4 $\times$ AMD EPYC 7K62 48-Core Processor}\\
        \multicolumn{2}{|c|}{CPU Memory} & \multicolumn{4}{c|}{32 $\times$ 32GB 2933MHz DDR4} \\
        \hline
        \multicolumn{2}{|c|}{GPU} & \multicolumn{4}{c|}{8 $\times$ NVIDIA Tesla A100 Tensor Core GPU}\\
        \multicolumn{2}{|c|}{GPU Memory} & \multicolumn{4}{c|}{8 $\times$ 40GB HBM2} \\
        \hline
        \multicolumn{2}{|c|}{SSD} & \multicolumn{4}{c|}{11TB} \\
        \hline
        \end{tabular}
    \end{center}
\label{tab:hard_ware}
\end{table}

\section{Experimental Evaluation}
\label{sec:experiment}

\begin{table}[t]
    \small
    \caption{Models for Evaluation.}
    \vspace{-3mm}
    \begin{center}
        \begin{tabular}{|c|ccccc|}
        \hline 
        \textbf{Model} & \textbf{\#Layer}  &  \textbf{\#Head} & $\mathbf{d_{Model}}$ & $\mathbf{d_{FFN}}$ & \textbf{\#Expert}\\ 
        \hline
        \hline
        GPT3-1.7B  & 24 & 24 & 2304 & 9216 &  -\\
        GPT3-13B   & 40 & 40 & 5140 & 20506 & -\\
        GPT3-28B   & 26 & 128 & 8192 & 32768 & -\\
        GPT3-30B   & 64 & 36 & 8192 & 32768 & -\\
        GPT3-55B   & 68 & 128 & 8192 & 32768 & -\\
        GPT3-120B   & 64 & 96 & 12288 & 49152 & -\\
        GPT3-175B   & 70 & 112 & 14336 & 57344 & -\\
        \hline
        T5-1.4B  & 16  & 16 & 1024 & 16384 & - \\
        T5-27B  & 28  & 64 & 4096 & 16384 & - \\
        T5-58B  & 60  & 64 & 4096 & 16384 & - \\
        T5-MoE-1.2T  & 16  & 16 & 1024 & 16384 & 2304\\
        \hline
        \end{tabular}
    \end{center}
\label{tab:model_structure}
\end{table}

\subsection{Experiment Setup}

\textbf{Machine environment.} 
Ours experiments are conducted on a production-grade GPU cluster in Tencent, where each server is equipped with 8 NVIDIA Tesla A100 GPUs and more details are shown in Table~\ref{tab:hard_ware}. 
GPUs are connected via NVLink-3.0 within a server, while servers are connected with RoCE using 16 NICs (16 $\times$ 12.5 GB/s in total). The PCIe bandwidth is 32GB/s while the SSD peak bandwidth is 3.5GB/s.

\textbf{Benchmarks.}
We conduct evaluations on three large-scale Transformer models, namely GPT-3~\cite{gpt3}, T5~\cite{t5}, and T5-MoE~\cite{fedus2021switch}, to validate the effectiveness and scalability of our proposed system. To achieve different model sizes, we experiment with varying numbers of Transformer blocks, hidden dimensions, and experts, and the specific model configurations are presented in Table~\ref{tab:model_structure}. We train all of these models using the mixed precision technique as introduced in Section~\ref{sec:background}, which stores the model states in FP32 while computes in BF16.

\textbf{Baselines.}  
Prior to \oursys{}, DeepSpeed~\cite{zerooffload} and Megatron-LM~\cite{megatron-lm}, due to their widespread adoption, are the two pre-training systems incorporated in the Taiji (as know as Angel) Machine Learning Platform of Tencent.
Therefore, we choose these two systems as our baselines to evaluate the effectiveness of \oursys{}. 
DeepSpeed is a heterogeneous training solution that currently achieves state-of-the-art performance, and we use the official examples to ensure a fair comparison. 
For Megatron-LM, we manually search the best parallelism strategy for each experimented model, which results in a hybrid parallelism solution that combines data parallelism, model parallelism, and pipeline parallelism.

We conduct a series of experiments to present the effectiveness and scalability of \oursys{} in training large-scale Transformer models.
Specifically, we evaluate the maximum model scale supported by our system on a single server in Section~\ref{sec:model_scale} and compare the throughput of different models on two GPU settings in Section~\ref{sec:throughput}. Additionally, we demonstrate the scalability of our system on hundreds of GPUs for training both billion-scale dense models and trillion-scale sparse models in Section~\ref{sec:scalability}.
Furthermore, in Section~\ref{sec:async}, we evaluate the convergence performance of our proposed \textit{Lock-Free Updating Mechanism} to further validate the effectiveness of our system when introducing SSD storage to scale the model size to an extreme level. 
It is worth noting that, except for experiments in Section~\ref{sec:async}, which utilizes SSD storage for training, all other sections utilize the memory of CPUs and GPUs by default.

\begin{table}[t]
    \small
    \caption{Max Supported Model Scale on a Single Server.}
    \vspace{-3mm}
    \begin{center}
        \begin{tabular}{|c|c|cccc|}
        \hline 
        \textbf{Model} & \textbf{System} & \textbf{\#Params} & \textbf{\#Batch} 
        & \textbf{GPU Mem} & \textbf{Samples/s}\\ 
        \hline
        \hline
        \multirow{4}{*}{GPT} & \multirow{2}{*}{DeepSpeed}  & 28B & 1 & 18 & 0.404 \\
        & & 28B & 36 & 40 & 7.61 \\
        \cline{2-6}
        & \multirow{3}{*}{AngelPTM}  & 28B & 38 & 39 & \textbf{10.99} \\
        && \textbf{55B} & 1 & 33 & 0.464 \\
        && \textbf{55B} & 10 & 40 & 3.34 \\
        \hline
        \multirow{4}{*}{T5} & \multirow{2}{*}{DeepSpeed}  & 27B & 1 & 20 & 0.317 \\
        & & 27B & 32 & 39 & 7.31 \\
        \cline{2-6}
        & \multirow{3}{*}{AngelPTM}  & 27B & 50 & 40 & \textbf{14.38} \\
        && \textbf{58B} & 1 & 38 & 0.432 \\
        && \textbf{58B} & 4 & 40 & 3.37 \\
        \hline
        \end{tabular}
    \end{center}
\label{tab:model_scale}
\end{table}
\vspace{-5mm}

\subsection{Model Scale}
\label{sec:model_scale}
We first conduct evaluations on a single server to test the maximum model size and corresponding maximum throughput that can be supported by \oursys{} and DeepSpeed, where we increase the number of transformer blocks and fix other model settings.
Both systems partition the model evenly across multiple GPUs using ZeRO-3~\cite{rajbhandari2020zero}, which enables efficient linear scaling for model size. 

Results are summarized in Table~\ref{tab:model_scale}.
For GPT models, we set the number of heads as 128, the embedding dimension as 8192, and the FFN hidden size as 32768. DeepSpeed can support a maximum model scale of 28B with 26 layers, while \oursys{} can further scale it up to 55B with 68 layers, which is a $96.4\%$ improvement. Note that despite each GPU having 22GB of memory available, DeepSpeed fails to scale to a larger model size. The reason is that since DeepSpeed statically partitions the model states across GPUs and CPUs, the maximum model scale will be limited by the CPU memory. 
In contrast, to fully exploit this available memory, \oursys{} uses the dynamic memory management that moves partial model states into GPU memory to achieve larger model scale. 
Regarding training efficiency, specifically the samples/s column in Table~\ref{tab:model_scale}, 
the maximum trhoughput of DeepSpeed is 7.61 samples/s, while that of \oursys{} is 10.99 samples/s, which is a $44\%$ improvement.
Furthermore, the training efficiency of \oursys{} for the largest supported GPT model, 55B, is 3.34 samples/s. These analyses can also be adapted to T5 models, where \oursys{} achieves $114.8\%$ improvement in terms of max model scale as well as $96.7\%$ improvement in terms of throughput performance.

In summary, compared with DeepSpeed, \oursys{} can (1) support larger scale of models using the same hardware resources and (2) achieve higher training efficiency for the same model. 

\subsection{Throughput}
\label{sec:throughput}

\begin{figure}[t]
    \centering
    \includegraphics[width=\linewidth]{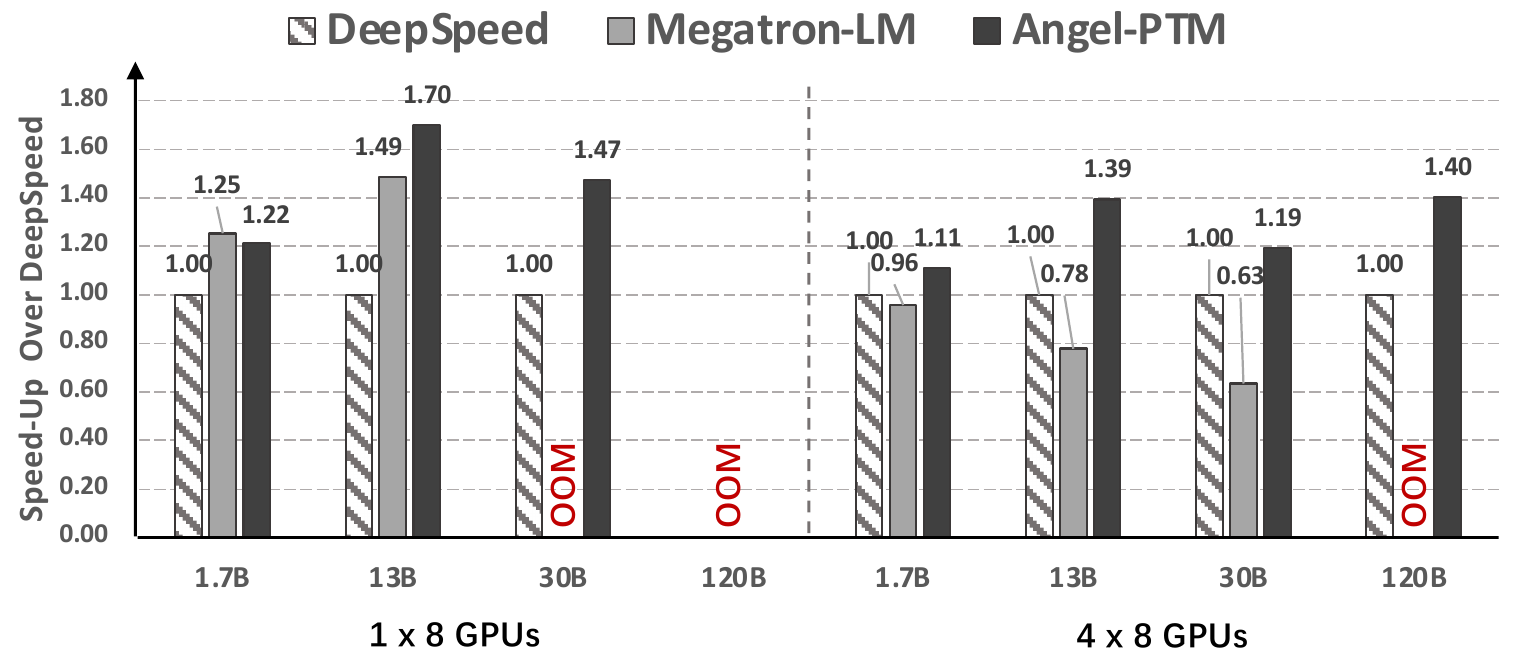}
    \caption{Compare \oursys{} with other famous pre-trained systems on a series of GPT models.}
    \label{fig:throughput}
    \vspace{-2mm}
\end{figure}

To verify the training efficiency of \oursys{}, we assess the throughput of each competitor. 
Specifically, we trained a series of GPT models with the maxmium batch size on 8 GPUs and 32 GPUs respectively, and the sizes of GPT models range from 1.7B to 120B.
To provide a clear comparison between the systems, we normalize the throughput of each system w.r.t. DeepSpeed. The results are presented in Figure~\ref{fig:throughput}.

In the configuration of 1x8 GPUs, \oursys{} consistently outperforms the other systems in terms of training efficiency, except for the 1.7B models. 
This is because the 1.7B model is small enough to be accommodated by a single GPU, and therefore the vanilla data parallelism (without ZeRO) achieves the best performance, which is also the strategy adopted by Megatron-LM. Since \oursys{} involves extra overhead on memory management even when memory movement is not needed, it runs slightly slower than Megatron-LM (a $2.4\%$ slowdown). 
On the 1.7B and 13B models, both \oursys{} and Megatron-LM outperform DeepSpeed. This is reasonable as DeepSpeed statically partitions models states between the CPU memory and the GPU memory, leading to redundant memory movements. 
However, as the model size increased to 30B, Megatron-LM fails with the out-of-memory error due to the limited GPU memory and \oursys{} still outperforms DeepSpeed because of our fine-grained life-time based scheduling method, which partitioned the model states to CPU on demand and scheduled the movements at the right time.

In the configuration of 4x8 GPUs, the performance of \oursys{} is still the best. 
With more GPUs, Megatron-LM is able to support the 30B model, while DeepSpeed and \oursys{} are further able to support the 120B model thanks to the ZeRO technique.
Both \oursys{} and DeepSpeed outperform Megatron-LM because they can train with larger micro batch sizes. 

In summary, the experimental results demonstrate that \oursys{} consistently outperforms both DeepSpeed and Megatron-LM in terms of throughput performance. Specifically, \oursys{} achieves an average of $35.4\%$ and up to $70\%$ improvement over the state-of-the-art hierarchical training system, DeepSpeed. It also outperforms the state-of-the-art hybrid parallelism training system, Megatron-LM, by an average of $38.9\%$ and up to $88.9\%$. 

\subsection{Scalability}
\label{sec:scalability}
To verify the scalability of \oursys{}, we conduct evaluations on two extremely large-scale Transformer models, which are GPT3-175B and T5-MoE-1.2T.

\textbf{Evaluations on GPT3-175B.} The GPT3-175B model is first proposed by OpenAI~\cite{gpt3}, which is trained on an enormous amount of text data and perfroms a range of NLP tasks, such as language translation and question answering.
It also acts as the foundation model for ChatGPT~\cite{chatgpt}, which is a groundbreaking achievement in the field of artificial intelligence. Therefore, it is crucial to verify our system's ability and scalability to support this model.

\begin{figure}[t]
    \centering
    \includegraphics[width=0.65\linewidth]{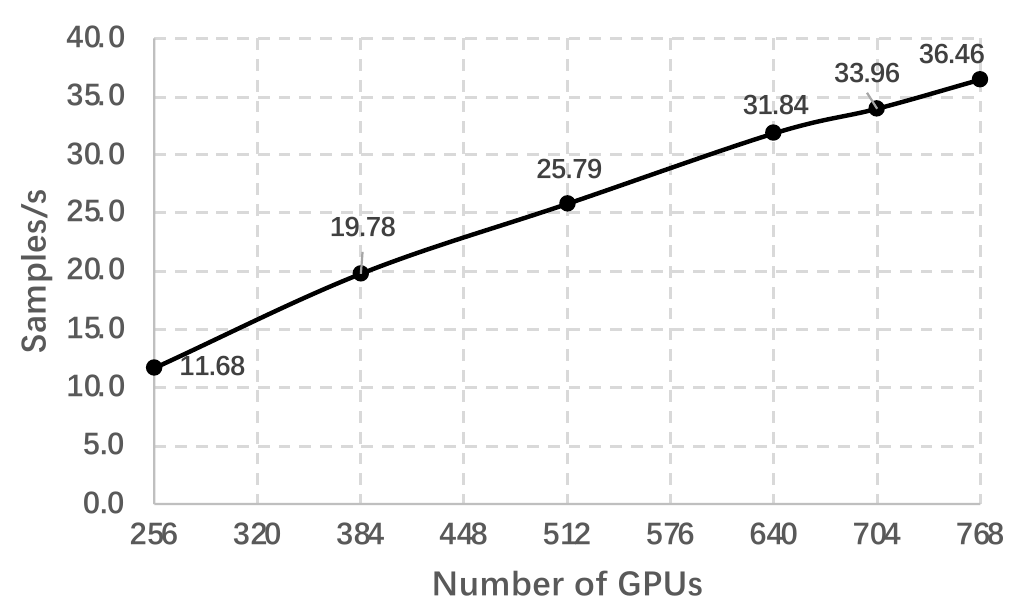}
    \caption{Scalability on training GPT3 models.}
    \label{fig:scalability_gpt3}
    \vspace{-3mm}
\end{figure}

The configuration of GPT-175B is detailed in Table~\ref{tab:model_structure} and we illustrate the throughput of training this model on different number of GPUs in Figure~\ref{fig:scalability_gpt3}. 
Our results demonstrate that \oursys{} achieves super-linear scalability when training the GPT-175B model. We observe a throughput of 11.68 samples/s on 32 nodes (256 GPUs), which increases to 36.46 samples/s on 96 nodes (768 GPUs), resulting in a $3.12\times$ speed-up. 
As the number of GPUs increases, the model states are distributed across more GPUs. This allows us to increase the global batch size, which in turn fully utilizes the available GPU memory. Moreover, the optimizer updating process is parallelized across more CPUs, and data movements are parallelized across more PCIes. These factors attribute to the super-linear scalability and higher training throughput.

The results indicate that our system can take full advantage of the increasing number of GPUs and achieve efficient parallelization of data movements on PCIe, which is critical for training large models in a timely and cost-effective manner.

\begin{figure}[h]
    \centering
    \includegraphics[width=0.65\linewidth]{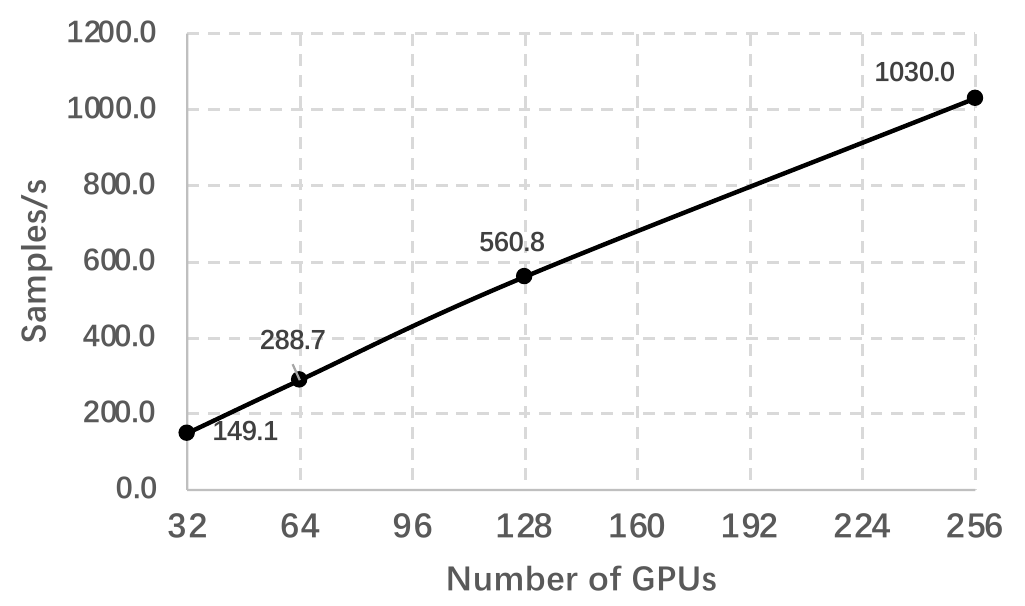}
    \caption{Scalability on training T5-MoE models.}
    \label{fig:scalability_T5_MoE}
     \vspace{-2mm}
\end{figure}

\textbf{Evaluations on T5-MoE-1.2T.} 
The T5-MoE model is another well-known line of large models that employs a sparse MoE architecture and was first proposed by Google as Switch-Transformer~\cite{GShard, fedus2021switch}. MoE models can significantly reduce training cost by routing input data to only a small number of expert networks. \oursys{} trained T5-MoE models using expert parallelism~\cite{nie2023flexmoe}, where expert parameters within an MoE layer are sharded among all GPUs while non-MoE parameters are duplicated.

The T5-MoE-1.2T model has 2304 experts per MoE layer, and the number of experts per GPU per MoE layer is fixed at 9 to achieve different model sizes when varying the number of GPUs. For example, the T5-MoE model trained on 128 GPUs has 1152 experts per MoE layer. The detailed configuration is presented in Table~\ref{tab:model_structure}.
We summarize the results in Figure~\ref{fig:scalability_T5_MoE}, which indicates that \oursys{} has near-linear scalability when training the T5-MoE model. 
With more servers involved in training, more input data will be feed into the all-to-all communication of the MoE layer, which can result in throughput degradation. Thus, the scalability on T5-MoE-1.2T is lower than that on GPT3-175B.

\subsection{Advancing Support for Extreme Model Scale}
\label{sec:async}
As noted by \citet{scalinglaws}, larger models tend to outperform smaller ones, which has motivated many researchers to increase the size of their models continuously. In this study, we evaluate the ability of \oursys{} in supporting extreme-scale models, using a scaled-up version of the T5-MoE model with 10T parameters.

We conduct pre-training of T5-MoE-1T and T5-MoE-10T based on an industrial text dataset. To support extreme model scales, we utilize the SSD storage into training. 
The experimental results are presented in Table~\ref{tab:async}, including training efficiency in samples/s and model quality in terms of validation loss. 
\oursys{} achieves a throughput of $37.26$ samples/s using 64 GPUs on the T5-MoE-1T model. When we scale up the model to 10T by increasing the number of experts, \oursys{} achieves a throughput of $317.72$ samples/s with 64 GPUs, demonstrating the near-linear scalability. 

Moreover, when introducing SSD training, the I/O bandwidth of SSD significantly affects the speed of CPU parameter updates, slowing down the overall training speed. By leveraging the \textit{Lock-Free Updating Mechnaism} in Section~\ref{sec:async} to perform asynchronous updates between CPUs and GPUs, \oursys{} significantly improves the overall training throughput. To be specific, for the T5-MoE-10T model, the training throughput increased from $317.82$ samples/s to $962.31$ samples/s when the lock-free mechanism is enabled, achieving a speedup of $2.96\times$. Meanwhile, experimental results on the validation loss verify that this mechanism has little impact to the model quality.

\begin{table}[t]
    \small
    \caption{Evaluation of the large-scale T5-MoE model training with SSD .}
    \begin{center}
        \begin{tabular}{|c|cccc|}
        \hline 
        \textbf{System} & \textbf{\#Params} & \textbf{\#GPUs} 
        & \textbf{Samples/s} & \textbf{Valid Loss$\downarrow$}\\ 
        \hline
        \multirow{2}{*}{AngelPTM}  & 1T & 64 & 37.26  & 1.124 \\
        & 10T & 576 & 317.82 & \textbf{0.853} \\
        \cline{2-5}
        + \textit{Lock-Free} & 10T & 576 & \textbf{942.31} & 0.861 \\
        \hline
        \end{tabular}
    \end{center}
\label{tab:async}
\end{table}

\section{Related Work}
\label{sec:related_work}
\textbf{Distributed Training System.}
Many well-known systems has been designed and implemented for large-scale model training. DeepSpeed proposed the ZeRO optimization on data parallelism, which evenly partitions model states across all devices with different optimization levels, including optimizer states
(e.g., 32-bit parameter, the first and second moments of Adam~\cite{kingma2014adam}) 
in stage 1, gradients in stage 2 and 16-bit parameters in stage 3~\cite{rajbhandari2022deepspeed}. However, this approach introduces extra communication to obtain the fully updated model parameters, 
leading to more time overheads for higher optimization levels, although they result in more memory reductions.
Megatron-LM proposed a novel approach called tensor-parallelism, which is both easy to implement and efficient in executing on highly connected GPUs such as A100 with NVLink-3.0~\cite{megatron-lm}. 
Galvatron~\cite{miao2022galvatron} proposed an efficient algorithm for finding the optimal strategy to combine data-parallelism, model-parallelism, pipeline-parallelism and ZeRO optimization together for efficiently training large model.
ZeRO can not be used for scaling large models due to the limited GPU memory and the model parallelism techniques, including Megatron-LM and Galvatron, are difficult to deploy in industry settings due to their complexity. \oursys{} adresses these challenges by integrating ZeRO with hierarchical memory, which achieves flexiblity as well as good scalability.

\textbf{Heterogeneous Training System.} 
With the evolution of deep learning, the type of models being trained has also evolved from early CNN-based models~\cite{resnet} to current transformer-based models~\cite{bert, gpt2, gpt3, t5}.
For CNN-based models, the memory usage is mainly occupied by activations, and research is mainly focused on optimizing single-GPU memory.
~\citet{wang2018superneurons} proposed evicting some activations by either recomputing them or offloading them to CPU based on cost analysis.
~\citet{nie2022tsplit} proposed the tensor-splitting optimization for fine-grained memory management, which can help break memory bottlenecks and lead to more efficient execution plans. 
Transformer-based models have high memory usage due to their large model states, and research efforts have focused on optimizing GPU memory for distributed training.
~\citet{zerooffload} proposed offloading optimizer computations and model states to the CPU to save GPU memory, and further combined this with a unique optimal offloading strategy and ZeRO-powered data parallelism.
\citet{zeroinfinity} introduced SSD storage into training and proposed a bandwidth-centric partitioning algorithm to distribute the model among all devices.
~\citet{patrickstar} dynamically managed the model states during training via a chunk-based memory manager. Different from existing systems, \oursys{} employs a fine-grained memory management via \emph{Page} to reduce the memory fragments and improve the training efficiency.

\textbf{GPU Memory Management.}
Many memory management optimizations have been proposed to pre-allocate most GPU memory and then manage the memory themselves, including paging~\cite{DBLP:conf/micro/Ausavarungnirun17}, replacement caching~\cite{DBLP:journals/pvldb/WangZYMLD014} and memory pool~\cite{DBLP:journals/corr/ZhangEfficient}. 
Mosaic~\cite{DBLP:conf/micro/Ausavarungnirun17} provided application-transparent support for multiple page sizes to page-in and page-out. 
MultiQx-GPU~\cite{DBLP:journals/pvldb/WangZYMLD014} designed a cost-driven replacement policy for efficient executions of concurrent queries in GPU databases.
\citet{DBLP:journals/corr/ZhangEfficient} proposed a memory pool, \textit{CNMeM}, which utilizes lifetime semantics to reduce memory fragments and designed a heuristic algorithm to simplify the optimization problem.
However, studies on paging, replacement caching, and unified memory address are not designed for deep learning training and do not utilize the special nature of tensor access patterns, while others about memory pool do not consider CPU memory. \oursys{} utilized the life-time information to improve overall performance, which reduced the fragements and improve the overlap between different resources. 


\section{Conclusion}
\label{sec:conc}
This work introduced \oursys{}, an easy-to-use and highly-efficienct deep learning systems for pre-training and fine-tuning tasks in Tencent. To be user-friendly and seamlessly scalable, we designed \oursys{} with the basis of data parallelism, parameter sharding, and hierarchical memory. To fully utilize the memory and bandwidth during training, a \emph{Page} abstraction was introduced to enable the fine-grained memory management along with a unified scheduling method was proposed to holistically manage the key operations during training. Moreover, we integrated the SSD storage to boost the pre-trained models to extreme scale and developed a lock-free updating mechanism to address the SSD I/O bandwidth bottleneck. Empirical results showed that \oursys{} outperforms existing systems in terms of both maximum supported model scale and training throughput.


\nocite{tinyscript,sketchml,vfboost,celu_vfl}
\bibliographystyle{ACM-Reference-Format}
\bibliography{sample}

\end{document}